\documentclass[sigconf]{acmart}

\usepackage{times}
\usepackage{latexsym}
\usepackage{caption} 
\usepackage{xcolor}
\usepackage{amsmath}
\usepackage{amsthm}
\usepackage{graphicx}
\usepackage{subcaption}
\usepackage{tcolorbox}
\usepackage{booktabs}
\usepackage{multirow}
\usepackage{color}
\usepackage{multicol}
\usepackage{algorithm}
\usepackage{algpseudocode}
\AtBeginDocument{%
  \providecommand\BibTeX{{%
    \normalfont B\kern-0.5em{\scshape i\kern-0.25em b}\kern-0.8em\TeX}}}

\setcopyright{acmcopyright}
\copyrightyear{2023}
\acmYear{2023}
\acmDOI{XXXXXXX.XXXXXXX}

\acmConference[KDD'23]{}{August 6--August 10,
	2023}{Long Beach, CA}
\acmPrice{15.00}
\acmISBN{978-1-4503-XXXX-X/18/06}
\acmSubmissionID{}

\begin{document}

\title{FactReranker: Fact-guided Reranker for Faithful Radiology Report Summarization}
 \author{Qianqian Xie}
\email{qix4002@med.cornell.edu}
\orcid{0000-0002-9588-7454}
\affiliation{%
  \institution{Department of Population Health Science,\\
  Weill Cornell Medicine,
   Cornell University
  }
  \city{New York}
  \country{USA}
  \postcode{ NY 10065}
}
\author{Jiayu Zhou}
\email{jiayuz@msu.edu}
\affiliation{%
  \institution{Michigan State University
  }
  \city{East Lansing}
  \country{USA}
}
\author{Yifan Peng}
\email{yip4002@med.cornell.edu}
\author{Fei Wang}
\email{few2001@med.cornell.edu}
\affiliation{%
  \institution{Department of Population Health Science,\\
  Weill Cornell Medicine,
   Cornell University
  }
  \city{New York}
  \country{USA}
  \postcode{ NY 10065}
}
\renewcommand{\shortauthors}{Qianqian and Jiayu, et al.}

\begin{abstract}
Automatic radiology report summarization is a crucial clinical task, whose key challenge is to maintain factual accuracy between produced summaries and ground truth radiology findings. Existing research adopts reinforcement learning to directly optimize factual consistency metrics such as CheXBert or RadGraph score. However, their decoding method using greedy search or beam search considers no factual consistency when picking the optimal candidate, leading to limited factual consistency improvement. To address it, we propose a novel second-stage summarizing approach FactReranker, the first attempt that learns to choose the best summary from all candidates based on their estimated factual consistency score. We propose to extract medical facts of the input medical report, its gold summary, and candidate summaries based on the RadGraph schema and design the fact-guided reranker to efficiently incorporate the extracted medical facts for selecting the optimal summary. We decompose the fact-guided reranker into the factual knowledge graph generation and the factual scorer, which allows the reranker to model the mapping between the medical facts of the input text and its gold summary, thus can select the optimal summary even the gold summary can't be observed during inference. We also present a fact-based ranking metric (RadMRR) for measuring the ability of the reranker on selecting factual consistent candidates. Experimental results on two benchmark datasets demonstrate the superiority of our method in generating summaries with higher factual consistency scores when compared with existing methods.
\end{abstract}

\ccsdesc[500]{Information systems~Summarization}
\ccsdesc[500]{Computing methodologies~Natural language generation}
\ccsdesc[500]{Applied computing~Bioinformatics}

\keywords{Factual consistency, radiology reports summarization, medical knowledge graph}

\maketitle

\section{Introduction}
\label{sec:intro}
\begin{figure}[!hbt]
    \centering
    \includegraphics[width=0.5\textwidth]{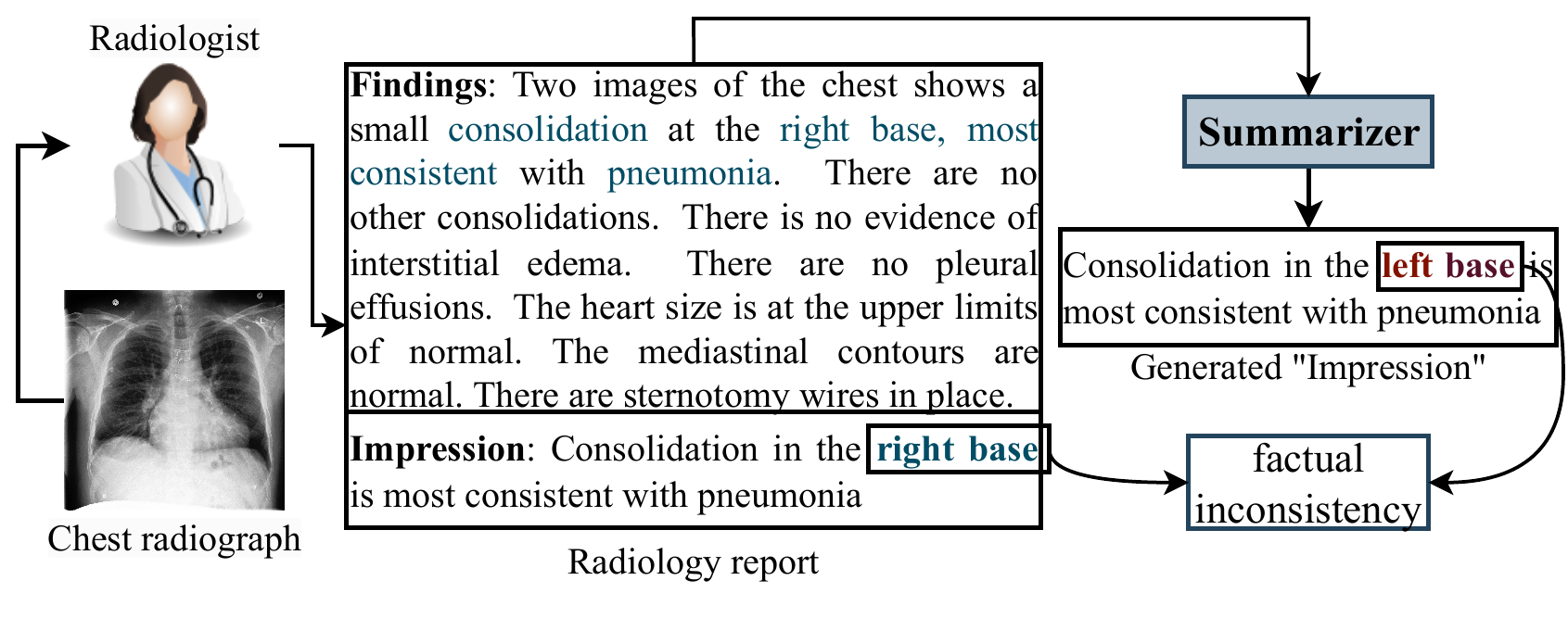}
    \caption{The example of automatic radiology report summarization and the factual inconsistency problem.}
    \label{fig:example}
\end{figure}

Radiology reports are an essential component of patient care, as they provide valuable information to physicians and other healthcare providers to help guide treatment decisions.
They usually include the “Findings” section illustrating detailed medical observations and findings of medical images, and the succinct “Impression” section preserving the summary of most preeminent observations which is the most important information for clinical decision-making.
Automatic radiology report summarization~\cite{DBLP:conf/acl/ZhangMTML20,zhang-etal-2018-learning-summarize,hu2021word} seeks to condense the "Findings" section of radiology reports into the succinct "Impression" section using the text summarization techniques, as shown in the Figure \ref{fig:example}.
It is a key clinical task with significant application value for reducing the burden of clinicians and enhancing their efficiency~\cite{wang2021pre}.
Maintaining the factual correctness between generated summaries and the ground truth radiology findings is of utmost importance for automated radiology report summarization.
In application, any errors or inaccuracies in the generated impressions could have significant implications for the patient's diagnosis and treatment.
To date, there have been few studies~\cite{DBLP:conf/acl/ZhangMTML20,Delbrouck2022TowardET} that focus on optimizing the factual correctness of generated summaries with reinforcement learning (RL).
\citet{DBLP:conf/acl/ZhangMTML20} developed a CheXBert~\cite{DBLP:journals/corr/abs-2004-09167} based metric to assess the factual correctness of clinical information in generated summaries and employ an RL based method to optimize the factual correctness objective based on the metric.
\citet{Delbrouck2022TowardET} further developed a RadGraph~\cite{DBLP:conf/nips/JainASTDBC0LNLR21} based metric, which utilizes a broader range of clinical information from radiology reports than the ChexBERT-based score and is more aligned with radiologists.

However, there exists a significant gap between the oracle performance of these models and their actual outputs, due to no \textbf{factual guidance} during the decoding process.
The vast majority of these methods adopt decoding strategies such as greedy search, beam search, and diverse beam search~\cite{vijayakumar2016diverse} et al, which does not factor in factual consistency as a criterion for selecting the best candidate summary from $k$ generated summaries.
These decoding methods generally maintain a list of top-$k$ best candidates and output the best one which is sorted according to log-probability.
Nevertheless, the left-out $k-1$ candidates can contain more correct facts.
To highlight the phenomenon, in Table \ref{tab:oracle}, we report the \textit{oracle} performances (the highest scores over all candidates) of three popular decoding methods with different metrics on the benchmark dataset MIMIC-CXR~\cite{Johnson2019MIMICCXRAL}.
We can see that the oracle performance when selecting the best candidate summary with the highest RadGraph score is significantly higher ($+16\%$) than existing SOTA methods~\cite{Hu2022GraphEC,Delbrouck2022TowardET} and the base model $\text{BART}_{BS}$.
It indicates a significant opportunity to improve the factual consistency of generated summaries by identifying better candidates.
\begin{table}[ht]
    \centering
    \small
    \begin{tabular}{cccccc}
        \hline
        Decoding strategy & R-1&R-2&R-L&$F_1$CheXbert&RadGraph\\
        \hline
        $\text{BART}_{BS}$&51.25&35.53&47.10&72.60&45.97\\
         $\text{BART}_{BS}$&51.25&35.53&47.10&72.60&45.97\\
         \citet{Hu2022GraphEC}& 51.02&35.21&46.65&70.73&45.23\\
         \citet{Delbrouck2022TowardET}&51.96&35.65&47.10&74.86&48.23\\
        \hline
        Beam search&55.09&39.59&50.69&74.63&64.35\\
        Diverse beam search&54.39&38.98&50.07&74.94&63.37\\
        Top-k sampling&53.03&37.21&47.58&73.04&49.62\\
        \hline
    \end{tabular}
    \caption{Oracle score for the BART~\cite{lewis2020bart} backbone model with different decoding strategies on different evaluation metrics on the benchmark dataset MIMIC-CXR~\cite{Johnson2019MIMICCXRAL}. The last three columns show the oracle performance of the base model BART with different decoding strategies. In all decoding methods, we keep 10 candidate summaries. R-1/2/L means the ROUGE-1/2/L score~\cite{Lin2004ROUGEAP}. $F_1$CheXbert and RadGraph score are two metrics for evaluating the factual consistency.}
    \label{tab:oracle}
\end{table}
\begin{figure}[!hbt]
    \centering
    \includegraphics[width=0.48\textwidth]{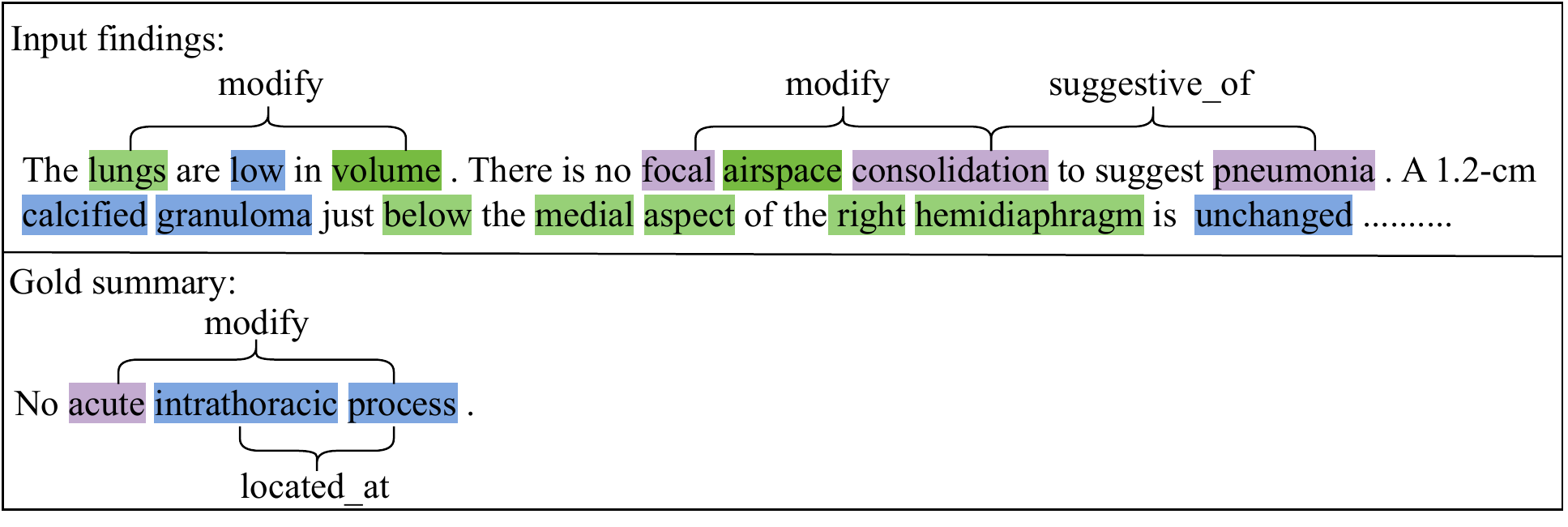}
    \caption{The example of the medical entities and relations in the input findings and its gold summary. We use different color to identify different type of entities. We also show the relations between entities such as "modify" and "suggestive\_of".}
    \label{fig:fc}
\end{figure}

To address these issues, in this paper, we propose a \textit{second-stage} summarization method \textbf{FactReranker}, which aims to enhance the factual correctness of generated summaries by learning to select the optimal summary from all candidates according to their factual correctness scores.
However, training a robust and effective reranker in the radiology report summarization scenario is a significant challenge for several reasons.
Firstly, different from general metrics such as ROUGE~\cite{Lin2004ROUGEAP} and Bertscore~\cite{DBLP:conf/iclr/ZhangKWWA20} that only consider the semantic similarity between the candidate and the gold summary, the factual consistency metric such as the RadGraph based metric~\cite{Delbrouck2022TowardET} calculates the overlapping between the medical entities and relations of the candidate and the gold summary.
This means the reranker should have the ability to identify and evaluate the medical facts (medical entities and relations) in the candidate summaries, to precisely select the best summary with the highest factual consistency score.
Secondly, another challenge for the reranker to effectively select the best summary that aligns with the gold summary in medical facts, arises from the significant gap of medical facts between the original input and the gold summary.
The reranker only has access to the information in the candidate summaries and the input radiology reports during inference, while the gold summary can't be observed.
Yet there is often a significant difference in the medical facts between the original input and the gold summary, i.e., as shown in Figure \ref{fig:fc}, the input findings contain multiple entities and relations that are not mentioned in the gold summary. 

\begin{figure}[!hbt]
    \centering
    \includegraphics[width=0.48\textwidth]{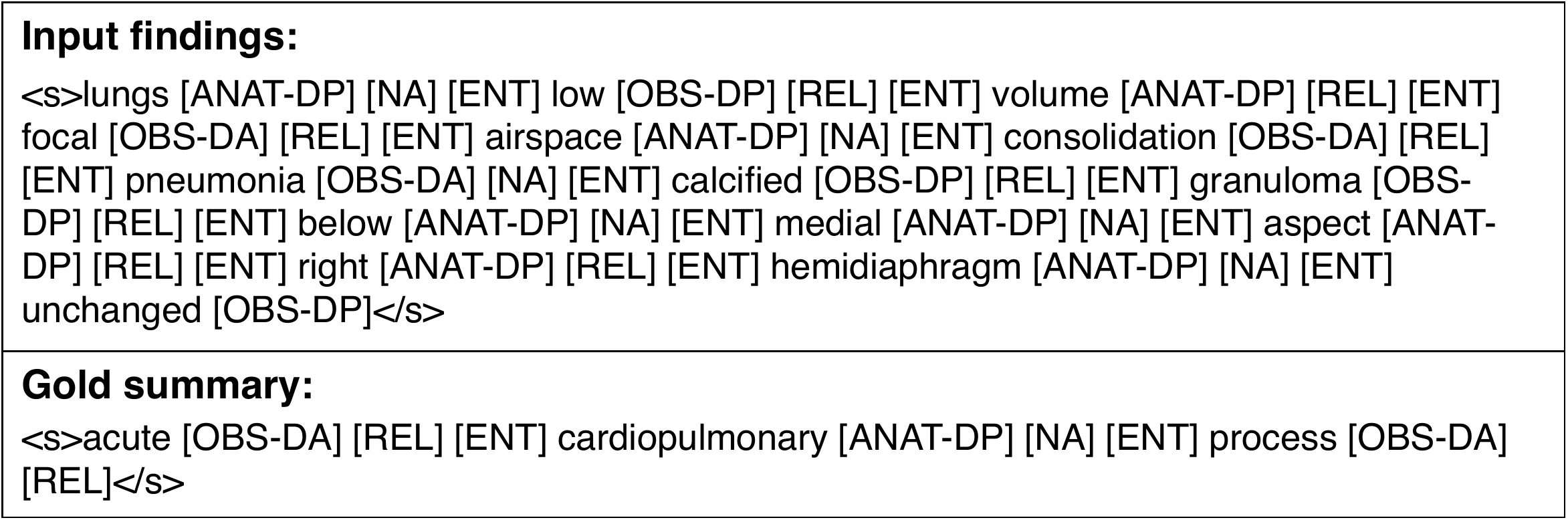}
    \caption{The example of the linearization of the RadGraph triplets in both the factual knowledge graph of input findings and its gold summary. We use the four types of entity same as the definitions in RadGraph schema and two simplified relation types as [REL](has any relation) and [NA](no relation), which is the same as the settings of the RadGraph partial scorer~\cite{Delbrouck2022TowardET}.}
    \label{fig:sequence-example}
\end{figure}

To address the aforementioned two challenges, we propose to extract medical facts from the input medical report, candidate summaries, and the gold summary based on the RadGraph schema, and develop the fact-guided reranker with the factual knowledge graph generation and the factual scorer based on extracted factual knowledge graphs, to select the optimal summary by explicitly capturing medical facts and the mapping of medical facts between the input texts and its gold summary.
The RadGraph~\cite{DBLP:conf/nips/JainASTDBC0LNLR21} is a dataset containing annotated clinical entities and relations of radiology reports based on the information extraction schema designed for structure radiology reports.
We generate a pool of candidates based on the backbone model with different decoding strategies in the first stage, and extract the factual knowledge graphs preserving medical facts including medical entities and relations of the input medical report, all candidates and the gold summary, based on the RadGraph schema.
We further develop the second stage fact-guided reranker with the factual knowledge graph generation and the factual scorer based on extracted factual knowledge graphs, that explicitly models the mapping between the medical facts of the gold summary and that of the input texts, enabling to select the optimal summary that factually aligns with the gold summary only based on medical facts of the input text, when the medical factual information of the gold summary can not be accessed during the inference.
To model the mapping between the medical facts of the gold summary and that of the input texts, the factual knowledge graph generation is to generate the factual knowledge graph of the target summary given the original input text and its factual knowledge graph.
Inspired by the generative knowledge graph construction~\cite{Zhang2021ContrastiveIE,Ye2020ContrastiveTE}, we build the factual knowledge graph generation as the sequence-to-sequence problem, which takes input as the linearization of RadGraph triplets in the factual knowledge graph of the input texts and outputs the linearization of expected RadGraph triplets in that of the target summary, as shown in Figure \ref{fig:sequence-example}.
Specifically, we design a new linearization for RadGraph triplets which starts with the entity token, followed by the special type tokens indicating the entity types and ended with the special relation tokens.
We also adopt a special token ([ENT]) to separate each triplet.
With the linearization, we can build a unified sequence for a factual knowledge graph that fully captures its structure information and label semantics.
The factual scorer thus can estimate the factual consistency score of candidate summaries based on the alignment between the generated factual knowledge graph of the target summary and that of candidates, and select the optimal summary with the highest alignment score.
Finally, for evaluating the ranking ability of fact-guided reranker, we design a fact-based ranking metric (\textbf{RadMRR}), that evaluates how well our method can select the factual consistent summary from all candidates according to the inverse of the rank of the summary with the highest factuality score.
Our contributions are as follows:
\begin{enumerate}
    \item We propose the FactReranker, a novel second-stage summarization framework for faithful radiology report summarization, by learning to select the optimal summary from all candidates, based on factual consistency and medical factual knowledge. To the best of our knowledge, this is the first second-state summarizer for factual enhanced radiology report summarization.
    \item We propose the novel fact-guided reranker consisting of the factual knowledge graph generation and factual scorer, which can select the best summary based on explicitly capturing medical factual knowledge and modeling the mapping of medical facts between the input texts and its gold summary. We design the new linearization of the knowledge factual graph which captures the structural information and label semantics, and enables us to model the factual knowledge graph generation as the sequence-to-sequence problem. We also present a fact-based ranking metric (RadMRR) for measuring the ability of the reranker on selecting factual consistent candidates.
    \item Experimental results over two benchmark radiology report datasets demonstrate the superiority of our method by improving the factual consistency score of generated summaries when compared with existing RL based summarization methods. Our method improves the RagGraph Score by 4.84\%, ROUGE-1 by 3.98\%, ROUGE-2 by 4.98\%, ROUGE-L by 4.75\%, and $F_1$CheXbert by 1.5\% on the MIMIC-CXR dataset, achieving the new SOTA.
\end{enumerate}

\section{Related Work}
\subsection{Radiology Report Summarization}
In recent years, a growing body of research has focused on automatic radiology report summarization.
\citet{zhang-etal-2018-learning-summarize} employed a bi-directional LSTM to perform sequence-to-sequence generation and reported that 30\% of generated summaries have factual errors.
\citet{DBLP:conf/acl/ZhangMTML20} proposed a factual correctness metric $F_1$CheXbert which used CheXbert to assess the factual consistency of the generated impressions.
The metric was then leveraged as the reward of a reinforcement learning (RL)-based summarizer to guide the training.
\citet{hu2021word} extracted medical entities using the Biomedical and Clinical English Model Packages, which are further exploited via graph neural networks.
Based on it,~\citet{Hu2022GraphEC} adopted the contrastive learning based pre-training that masks the entities in the findings to fully embed the domain knowledge.
Most recently,~\citet{Delbrouck2022TowardET} developed a RadGraph-based factual consistency measurement, and RL-based method which utilizes the metric as the reward.
However, these methods have no factual guidance during the decoding process via beam search or greedy search and fail to explicitly model the domain knowledge in both input documents and target summary via reinforcement learning.
To the best of our knowledge, we are the first second-stage summarization approach for radiology report summarization, which aims to rerank the generated candidates based on their factual consistency with the explicitly predicted medical facts in the target summary.

\subsection{Second-stage Abstractive Summarization}
Numerous research has focused on exploring the second-stage abstractive summarization techniques. 
\citet{Dou2020GSumAG} guided the abstractive summarization using the salient sentences selected by extractive methods.
\citet{Sun2021AlleviatingEB} proposed to employ contrastive learning to improve the chance of generating high-quality summaries and reduce exposure bias.
Similarly, \citet{Xu2021SequenceLC} optimized the similarities between the source document, the gold summary, and the candidates to improve faithfulness via contrastive learning.
\citet{Liu2021RefSumRN} introduced RefSum, which incorporates a meta-system for selecting the optimal summary from all candidates.
Regarding the reranker based methods in the second stage, 
SimCLS~\cite{Liu2021SimCLSAS} created an evaluation model with the objective of directly optimizing the evaluation metric using contrastive learning.
Similarly, BRIO~\cite{Liu2022BRIOBO} reused the generation model as the evaluation model to rank the candidates with contrastive learning.
SummaReranker~\cite{Ravaut2022SummaRerankerAM} trained a multi-task learning framework based on mixture-of-experts to optimize multiple metrics like ROUGE concurrently.
SummaFusion~\cite{Ravaut2022TowardsSC} combined numerous candidates to provide a more concise summary.
Even though these methods have shown promising performance in abstractive summarization for directly optimizing evaluation metrics,
there have been no prior efforts to explore the second-stage method for radiology report summarization.
Moreover, it is challenging to directly apply these methods in the general domain to the radiology report summarization task,
since it requires modeling and predicting the domain knowledge for factual consistency metrics in the medical domain, unlike commonly used metrics such as ROUGE.

\subsection{Generative Knowledge Graph Construction}
Existing knowledge graph construction (KGC) relies on a series of tasks including named entity recognition, entity linking, relation extraction, and event extraction, which suffers from error propagation in the pipeline and limited extension on different tasks.
To address it, a series of studies based on a sequence-to-sequence framework was proposed.
\citet{Zeng2018ExtractingRF} was the first attempt that revisits the KGC as a sequence-to-sequence problem and designed a copy mechanism to generate knowledge triplets.
Despite following research in the copy mechanism~\cite{Zeng2019CopyMTLCM,Zeng2019LearningTE,Huang2021DocumentlevelEE,Giorgi2022ASA},
there were studies utilizing structural knowledge and label semantics as unified text sequences.
\citet{Lu2021Text2EventCS} designed a linearization of the extracted events for event extraction based on T5.
\citet{Lou2021MLBiNetAC} further proposed a Multi-Layer Bidirectional Network to capture document-level relationships and semantics.
\citet{Zhang2021ContrastiveIE} and \citet{Ye2020ContrastiveTE} proposed to generate a sequence of triplets separated by tags via contrastive learning.
Based on it, \citet{Cabot2021REBELRE} designed a new linearization method for triplets that considers the order of triplets according to the head entity and tail entity.
Inspired by these studies, in this paper, we revisit the factual knowledge graph generation as a sequence-to-sequence generation task and design a new linearization for triplets of the factual knowledge graph, which captures the structural information and label semantics embedded in the triplets.

\section{Methodology}
\begin{figure*}[!hbt]
    \centering
    \includegraphics[width=\textwidth]{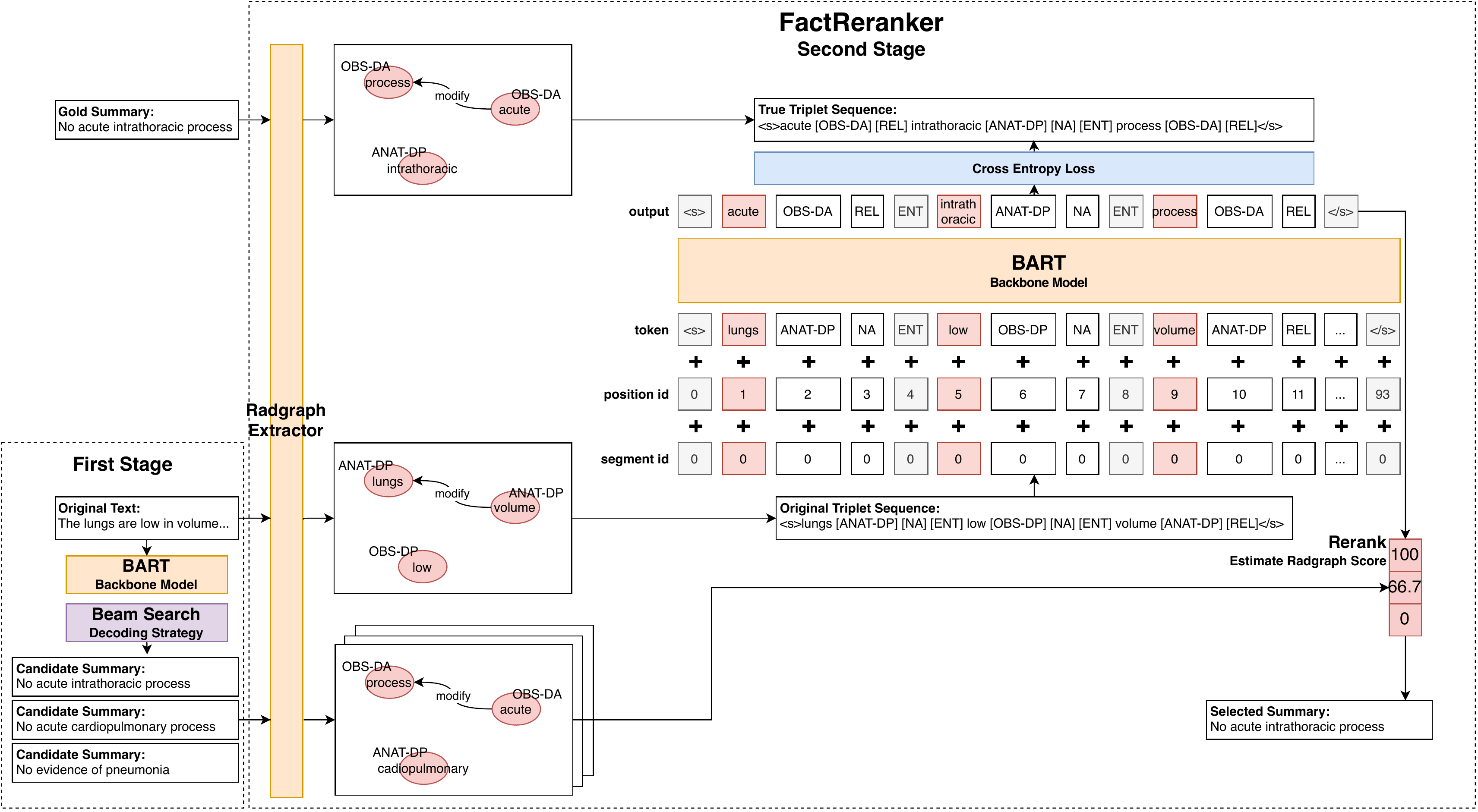}
    \caption{The overall framework.}
    \label{fig:fr}
\end{figure*}

In this section, we present the detail of our proposed framework, as shown in~\ref{fig:fr}.

\subsection{Framework of Two Stage Summarization Method}
Formally, given the input radiology findings with sequence of $N$ tokens $X = \{x_1, x_2, \dots, x_N\}$, the automatic summarization task on radiology reports is to generate the summary to best describe the findings of $X$, with a sequence $Y = \{y_1, \dots, y_i, \dots, y_L\}$, where $L$ is the length of impressions ($L << N$) and $y_i \in V$ is the generated token of the vocabulary $V$.
Consequently, the generation can be modelled as the conditional sequence to sequence generation problem:
\begin{equation}
    p(Y|X) = \prod^{L}_{i=1} p(y_i|y_1,\dots,y_{i-1},X)
\end{equation}

\subsubsection{First-stage summarizer}
For the two-stage summarization method, we first train a base summarization model with the following objective in the first stage, that maximizes the log-likelihood of the training data:
\begin{equation}
    \theta^* = \arg \max_\theta \sum_{i=1}^L \log p(y_i|y_1,\dots,y_{i-1},X;\theta)
    \label{eq:1}
\end{equation}
where $\theta$ is the parameter set of the model.
In this paper, we employ the pre-trained encoder-decoder language model BART~\cite{lewis2020bart} which is pre-trained on the text generation task as our backbone model, and fine-tune it with the maximum likelihood estimation (MLE) loss as shown in Equation \ref{eq:1} on the training data used in the radiology report summarization task. 
For each document $X$, we then generate a set of candidate summaries with the trained base model:
\begin{equation}
    C = \text{BASE}_{\theta^*} (X,S)
\end{equation}
where $C = \{C_1, C_2, \dots, C_m\}$, $S$ means the decoding strategy.
\subsubsection{Second-stage reranker}
The reranker aims to rerank all candidate summaries generated by the first-stage summarizer and select the best candidate according to factual consistency.
Given a factual consistency metric $\mu$, we can calculate the corresponding score for each candidate $S_\mu = \{\mu(C_1), \mu(C_2), \dots, \mu(C_m)\}$, based on their associated gold summaries in the training data.
In this paper, we aim to train a reranker $f_{\theta_s}(S_\mu, X, C)$ parameterized by $\theta_s$ that can identify the best summary candidate $C^*_\mu$ with the highest factual consistency:
\begin{equation}
    C^*_\mu = {\arg \max}_{C_i \in C} \{\mu(C_1), \dots, \mu(C_m)\}
\end{equation}

\subsection{RadGraph Score}
\subsubsection{Choice of factual consistency evaluation metric}
Up to now, there are two proposed factual consistency metrics for radiologist reports including the $F_1$CheXBert~\cite{DBLP:journals/corr/abs-2004-09167} and RadGraph score~\cite{DBLP:conf/nips/JainASTDBC0LNLR21}.
The $F_1$CheXBert computes the $F_1$ similarity score between the indicator vectors of 14 medical pathologies that are extracted from the generated and golden summaries with the ChexBERT.
Similar to $F_1$CheXBert, the RadGraph score is an F-score metric that measures the factual alignment (medical entities and relations) between the candidates and the gold summaries.
However, it has been proved that the $F_1$CheXBert score fails to align with the radiologists~\cite{yu2022evaluating,DBLP:conf/fat/BoagKRWG21,DBLP:conf/eccv/AndersonFJG16}, and is limited to a small part of clinical facts which only covers 14 medical observations, resulting in the poor evaluation of diverse clinical facts in radiology reports.
Compared with the $F_1$CheXBert metric, the RadGraph score covers diverse clinical facts of radiology reports and has higher alignment with radiologists.
Therefore, we use the RadGraph score as the optimized metric.
\subsubsection{Extracting medical facts}
\label{subsec:emf}
\begin{figure}[!hbt]
    \centering
    \includegraphics[width=0.48\textwidth]{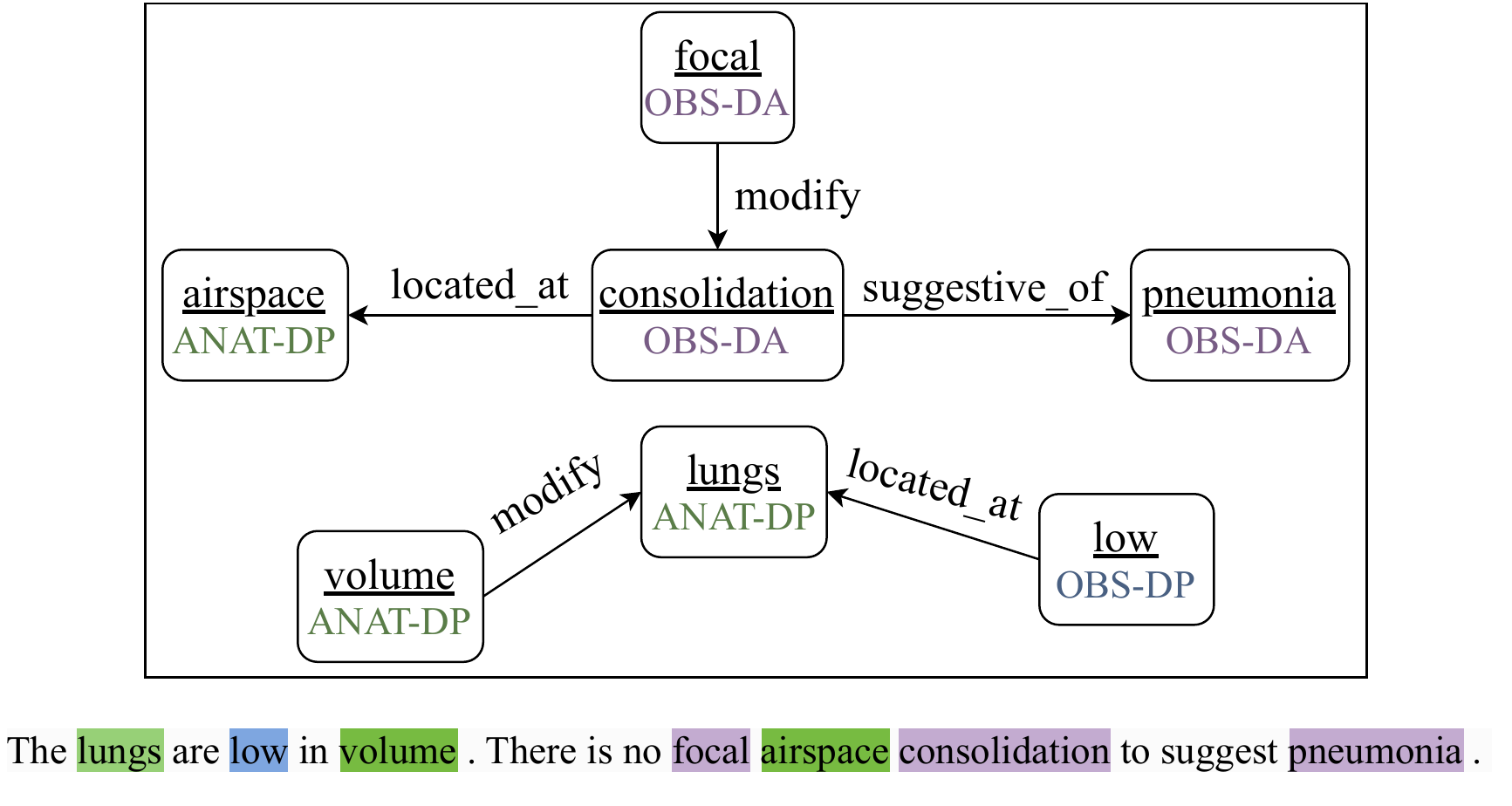}
    \caption{The example of the RadGraph of the input text extracted by the PubMedBERT model.}
    \label{fig:RadGraph}
\end{figure}

The RadGraph score of candidates calculates the overlap of their medical facts with that of the gold summary.
Therefore, we first use the PubMedBERT~\cite{gu2021domain} model\footnote{We use the pre-trained model checkpoint provided in: \url{https://physionet.org/content/RadGraph/1.0.0/}} pre-trained in annotated data based on the RadGraph schema, to extract medical facts including the medical entities and relations to build the RadGraph for the input report, its gold summary, and all candidates.
Formally, we consider the extracted RadGraph of the input text as a graph $\mathcal{G}(\mathcal{V}, \mathcal{E})$, with a set of entities as nodes $\mathcal{V} = \{v_1, v_2, \dots, v_{|V|}\}$ and the relations between nodes as edges $\mathcal{E} = \{e_1, e_2, \dots, e_{|E|}\}$, as shown in~\ref{fig:RadGraph}.
Each entity node in the RadGraph has a corresponding label $v_{i_L}$ which indicates its entity type.

\textbf{Entities}:
There are two types of entities in the RadGraph, including \textit{Anatomy} and \textit{Observation}.
For \textit{Observation}, there are also three different levels of uncertainty.
Consequently, we have four categories of entities: \textit{Anatomy (ANAT-DP)}, \textit{Observation: Definitely Present (OBS-DP)}, \textit{Observation:Uncertain (OBS-U)}, and \textit{Observation: Definitely Absent (OBS-DA)}.

\textbf{Relations}:
There are three types of relations between entities in the RadGraph, including the \textit{\{Suggestive\_Of\}} relation between two \textit{Observation} entities, the \textit{\{Located\_At\}} relation between the \textit{Observation} entity and the \textit{Anatomy} entity, and the \textit{\{Modify\}} relation between two \textit{Anatomy} entities.

To calculate the RadGraph score between candidates and the gold summary, we further transform the RadGraph into a set of triplets $T = \{(v_i, v_{i_L}, R)\}_{i=1:|V|}$, where $R = 1$ if there is any edges connected with $v_i$, otherwise $R = 0$.
Therefore, a triplet contains the entity, the entity type, and whether or not the entity has a relation.
Given the triplet sets for both the candidate $T_{C_i}$ and the gold summary $T_Y$, the RadGraph score is the harmonic mean of precision and recall between these two sets to measure the overlapping of medical facts between the candidate and the gold summary.
Therefore, we yield the RadGraph score for all candidates $C_i \in C$ for each document in the training data as:
\begin{equation}
    \mu(C_i, Y) = RadGraph(C_i, Y)
\end{equation}
which derives the $F_1$ score based on the triplet sets of the candidate $T_{C_i}$ and the gold summary $T_Y$.

\subsection{Fact-guided Reranker}
Based on the generated candidates and their RadGraph score calculated above, we aim to train the robust and effective reranker that can select the best candidate with the highest factual consistency score, namely the highest overlap of medical entities and relationships with that of the gold summary.
However, the gold summary is unobserved during the inference, and it is required to select the candidate summary with the trained reranker $f_{\theta_s}(C_i, X, T_{C_i}, T_{X})$, based on the observed candidate $C_i$ and the input text $X$ during the inference.
However, the medical facts of the input text has a significant gap with that of the gold summary, which leads to great challenge for training a effective reranker that can select the optimal summary during the inference.

Therefore, we propose the fact-guided reranker which can precisely rank the generated candidates with estimated factuality scores based on predicted medical facts in the target summary, when the ground truth medical facts of the gold summary can not be accessed during the inference.
To achieve this, we decompose the fact-guided reranker into two steps: 1) \textbf{factual knowledge graph generation:} it aims to generate the factual knowledge graph of the gold summary given the factual knowledge graph of the input text $X$, which models the mapping between the medical facts of the text input and its gold summary; and 2) \textbf{factual scorer:} it calculates the factuality score of each candidate $C_i$ according to the overlapping between its medical facts and that of generated factual knowledge graph of the gold summary.
Inspired by existing research in knowledge graph construction~\cite{Zhang2021ContrastiveIE,Ye2020ContrastiveTE}, we build the factual knowledge graph generation as a sequence-to-sequence task to better model the factual connections between entities and relations.
As shown in Figure~\ref{fig:sequence-example}, the input is the RadGraph triplets $T_X$ and the output is the RadGraph triplets $T_Y$ of the gold summary.

\subsubsection{Triplets Linearization}
We design a new linearization to express RadGraph triplets as a sequence of tokens, which can be modeled and generated by the backbone model such as the pre-trained language model to fully leverage its structural information and label semantics.
Given triplets of the input text $T_X = \{(v_i, v_{i_L}, R)\}_{i=1:|T_X|}$, the corresponding sequence would be:
\begin{equation}
    <s>v_0\ v_{0_L}\ R_0\ [ENT]\ v_1\ v_{1_L}\ R_1\ [ENT]\ \dots\ v_i\ v_{i_L}\ R_i</s>
\end{equation}
where \textit{<s>} and \textit{</s>} is the special token added before and after the generated sequence, $v_i$ is the entity token, $v_{i_L}$ is the entity label, and $R_i$ is the relation.
For the entity label $v_{i_L}$, we introduce four new special tokens corresponding to the four entity types in the RadGraph: [ANAT-DP] for \textit{Anatomy (ANAT-DP)}, [OBS-DP] for \textit{Observation: Definitely Present (OBS-DP)}, [OBS-U] for \textit{Observation:Uncertain (OBS-U)}, and [OBS-DA] for \textit{Observation: Definitely Absent (OBS-DA)}.
We also introduce two new special relation tokens for $R_i$, including [REL] for \textit{has\_relation} and [NA] for \textit{no\_relation}.
Additionally, we add the special token [ENT] between each triplet to indicate the end of the last triplet and the start of the next triplet.
Similar to \cite{Cabot2021REBELRE}, we sort the triplets by their first appearance in the text and linearize the triplets following that order.

\subsubsection{Factual Knowledge Graph Generation}
Given the input sequence as the linearization of the RadGraph triplets $T_X$, we adopt an autoregressive transformer model to generate the output sequence as the linearization of the expected triplets $T_Y$ in the gold summary.
Therefore, the task can also be deemed as a sequence-to-sequence task, which allows us to reuse and continue training the first-stage summarizer.
Specifically, the model consists of two major components:

\textbf{Input Encoder:} For entity token $v_i$, we tokenize the entity token, and append the introduced special tokens including entity types, relations, and separation in the tokenizer, which allows the model to not tokenize these tokens into sub-tokens. 
The final input representations are the summing of the corresponding token embedding, position embedding, and segment embedding.

\textbf{Generating Factual Knowledge Graph:}
We reuse the first stage summarization method based on BART and continual fine-tune it to generate the target sequence $T_Y$ of the gold summary based on the input representations on the training set.
Note that we adopt the partial casual masking in the decoder to perform auto-aggressive generation, which means tokens are only conditioned on leftward context to capture the order of RadGraph triplets in both the source document and target summary.
It is also notice that since we utilize three tokens in designed order to represent the triplet, i.e, the method should identify the token type of the previous entity token when generating the next token after the entity.
We then inference the RadGraph triplets of the target summary given the source document in the validation and test set based on the trained model.

\subsubsection{Factual Scorer}
Based on the generated sequence of the gold summary from the generation model, we can derive the corresponding triplet set $T'_Y$ of the gold summary.
We filter out the corrupted triplets in the generated sequence, i.e, whose entity type token and relation type token are not in the given categories or whose token numbers are not equal to three.
Consequently, given the estimated triplet set $T'_Y$ and extracted triplet set $T_{C_i}$ for each candidate summary $C_i$, we can calculate the estimated RadGraph score as:
\begin{equation}
    \mu'(C_i, Y) = RadGraph(T_{C_i}, T'_Y) =  2 * \frac{|T_{C_i} \cap T'_Y|}{|T_{C_i}| + |T'_Y|}
\end{equation}
Finally, our reranker can rank the generated candidate summaries according to their estimated RadGraph scores $\mu'(C_i, Y)$, and select the top-1 candidate as the final output during the inference.

\subsection{RadMRR}
\label{sec:radmrr}
Based on the factual consistency metric RadGraph score,
inspired by previous information retrieval research,
we  design a novel ranking metric named RadMRR which aims to evaluate the re-ranking performance of the reranker.
Given the observed candidate $C_i$, we calculate the reciprocal rank $\frac{1}{rank}$ of the position of the highest-ranked summary according to their RadGraph score.
For example, suppose that we have three candidates, the optimal candidate with the highest RadGraph score is ranked as the third item by the method, then the corresponding reciprocal rank would be $\frac{1}{3}$.
Given all examples, we average their reciprocal ranks as the mean reciprocal rank:
\begin{equation}
    RadMRR = \frac{1}{|C|} \sum_{C_i \in C} \frac{1}{rank_{C_i}}
\end{equation}
It is easy to compute and interpret and focus on the first summary of all candidates, which can assess the gap caused by inappropriate ranking.

\section{Experiments}
In this section, we conduct experiments to answer the following research questions about our method:
\textbf{RQ1:} Does the FactReranker improve the factual consistency of generated summaries for radiology report summarization?
\textbf{RQ2:} What component in FactReranker contributes to the improvement of factual consistency?
\textbf{RQ3:} How is the quality of the generated RadGraph for the gold summary via the factual knowledge graph generation model?

\subsection{Dataset}
Following previous methods~\cite{Hu2022GraphEC}, we conduct experiments on two widely-adopted radiology report datasets as \textbf{OpenI}~\cite{DemnerFushman2015PreparingAC} and \textbf{MIMIC-CXR}~\cite{Johnson2019MIMICCXRAL}.
OpenI contains 3,268 reports collected by Indiana University, while MIMIC-CXR consists of 124,577 reports.
We also filter out the reports as previous methods that have no findings, no impressions, findings with less than 10 words, or impressions with less than 2 words.
We use the same data splitting as previous methods~\cite{Hu2022GraphEC}, which divides OpenI into train/validation/test sets as 2400: 292: 576.
For MIMIC-CXR, we adopt the official split in the dataset released by~\citet{Johnson2019MIMICCXRAL}.
The statistics for two datasets are depicted in Table~\ref{tab:statistics}.
\begin{table}[htb]
\begin{tabular}{@{}l|l|l|l|l@{}}
\toprule
DATASET                    & TYPE      & TRAIN   & DEV   & TEST  \\ \midrule
\multirow{5}{*}{OpenI}     & REPORT \# & 2400    & 292   & 576   \\
                           & AVG.WF    & 37.89   & 37.77 & 37.98 \\
                           & AVG.SF    & 5.75    & 5.68  & 5.77  \\
                           & AVG.WI    & 10.43   & 11.22 & 10.61 \\
                           & AVG.SI    & 2.86    & 2.94  & 2.82  \\ \midrule
\multirow{5}{*}{MIMIC-CXR} & REPORT \# & 122,014 & 957   & 1606  \\
                           & AVG.WF    & 55.78   & 56.57 & 70.67 \\
                           & AVG.SF    & 6.50    & 6.51  & 7.28  \\
                           & AVG.WI    & 16.98   & 17.18 & 21.71 \\
                           & AVG.SI    & 3.02    & 3.04  & 3.49  \\\bottomrule
\end{tabular}
\caption{The statistics of OpenI and MIMIC-CXR, including the number of reports (REPORT \#), the average sentence number for both findings and impressions (WF \& SF), the average word number for both findings and impressions (WI \& SI).}
\label{tab:statistics}
\end{table}

\subsection{Baselines and Metrics}
We compare our methods to state-of-the-art summarization methods for radiology reports, including extractive summarization methods such as:
\begin{enumerate}
    \item \textbf{LEXRANK}~\cite{Erkan2004LexRankGL}, a commonly used graph-based ranking method,
    \item \textbf{TRANSFORMEREXT}~\cite{Liu2019TextSW}, a method using the Transformer~\cite{vaswani2017attention} network,
\end{enumerate}
and abstractive summarization approaches including:
\begin{enumerate}
\item \textbf{PGN}~\cite{see2017get}, an LSTM-based method using the pointer generate network and the copy mechanism for decoding,
    \item \textbf{TRASFROMERABS}~\cite{Liu2019TextSW}, a seq2seq method based on the Transformer network,
    \item \textbf{WGSUM (TRANS+GAT)}~\cite{hu2021word}, a method seq2seq method incorporating important words and their relations with the word graph,
    \item \textbf{GRAPHSUM}~\cite{Hu2022GraphEC}, a pre-trained language model based method incorporating extra knowledge, 
    \item \textbf{BART}~\cite{lewis2020bart}, the method fine-tuning the pre-trained language model BART,
    \item \textbf{RadGraph-RL}~\cite{Delbrouck2022TowardET}, the state-of-the-art factual guided summarization method using reinforcement learning to optimize the RadGraph score,
\end{enumerate}
Following previous methods, we adopt two types of metrics to evaluate performance.
First, for general summarization metrics, we employ three different ROUGE metrics including ROUGE-1 (R-1), ROUGE-2 (R-2), and ROUGE-L (R-L)~\cite{Lin2004ROUGEAP}.
Then, for factual consistency which is most important for radiology report summarization, we report both the RadGraph score~\cite{Delbrouck2022TowardET} and $F_1$CheXbert score~\cite{Zhang2019OptimizingTF}.
To be consistent with previous methods, we only report the RadGraph score and $F_1$CheXbert score on the MIMIC-CXR dataset, since the RadGraph score and $F_1$CheXbert score have been shown to be ineffective for the OpenI dataset.
The former is based on the overlap of domain entities and relations extracted via RadGraph, while the latter is the alignment of 14 abnormalities defined in CheXbert~\cite{Smit2020CheXbertCA}.

\subsection{Implementation Details}
In our experiments, we utilize BART implemented in the HuggingFace as the backbone model in the first stage, and use the parameter setting with learning rate=2e-5, training epoch=5, training/evaluation batch size=5, warmup steps=8000, and weight decay=0.01.
According to Table~\ref{tab:oracle}, we adopt the beam search as the decoding strategy for inference and generate 10 candidate summaries for each report.
We save the model checkpoint every 500 steps and select the best checkpoint based on the validation set.
For extracting the RadGraph, we use the model checkpoint based on the PubMedBERT\footnote{\url{https://physionet.org/content/RadGraph/1.0.0/}}, and the released code\footnote{\url{https://github.com/dwadden/dygiepp}}.
In the reranker, we use the best checkpoint in the first stage as the backbone method for triplet generation and set the learning rate to 1e-4, training epochs to 5, and dropout to 0.5.

\subsection{Main Results}
To answer \textbf{RQ1}, in Table~\ref{tab:main-results}, we present the performance of all methods on OpenI and MIMIC-CXR datasets on all metrics.
\begin{table*}[htb]
\begin{tabular}{@{}l|cll|cllll@{}}
\toprule
MODEL         & \multicolumn{3}{c|}{OpenI}           & \multicolumn{5}{c}{MIMIC-CXR}                                   \\ \midrule
METRIC        & R-1                  & R-2   & R-3 
& R-1                  & R-2   & R-3   & $F_1$CheXbert & RadGraph \\ \midrule
LEXRANK~\cite{Erkan2004LexRankGL}      & 14.63                & 4.42  & 14.06 
& 18.11                & 7.47  & 16.87 & -             & -        \\
TRANSEXT~\cite{Liu2019TextSW}      & 15.58                & 5.28  & 14.42
& 31.00               & 16.55 & 27.49 & -             & -        \\
PGN~\cite{see2017get}     & 63.71                & 54.23 & 63.38 
& 46.41                & 32.33 & 44.76 & -             & -        \\ \midrule
TRANSABS~\cite{Liu2019TextSW}&59.66&49.41&59.18&47.16&32.31&45.47&-        & - \\
WGSUM~\cite{hu2021word} & 61.63                & 50.98 & 61.73
& 48.37                & 33.34 & 46.68 & 69.15         & 44.20    \\
GRAPHSUM~\cite{Hu2022GraphEC}      & 64.97                & 55.59 & 64.45 
& 51.02                & 35.21 & 46.65 & 70.73         & 45.23    \\ \midrule
RadGraph-RL~\cite{Delbrouck2022TowardET}   & -                    & -     & -     
& 51.96                & 35.65 & 47.10 & 74.86         & 48.23    \\ 
BART~\cite{lewis2020bart}       &  63.68&54.28&62.96&50.50&35.27&46.61&71.00&45.10\\
\midrule
FactRanker          & 64.24&52.61&63.42&\textbf{55.94}&\textbf{40.63}&\textbf{51.85}&\textbf{76.36}&\textbf{53.17} \\ 
FactRanker(w/o RC)          & 63.69&54.28&62.97&50.51&35.28&46.62&71.00&45.08 \\ 
\bottomrule
\end{tabular}
\caption{The ROUGE F1, $F_1$CheXbert and RadGraph score of different methods on both OpenI and MIMIC-CXR datasets. Some results are referenced from ~\cite{hu2021word,Hu2022GraphEC,Delbrouck2022TowardET}}
\label{tab:main-results}
\end{table*}
\begin{table*}[htb]
\begin{tabular}{@{}l|l|l|llllll@{}}
\toprule
TYPE & TEXT & RadGraph & \begin{tabular}[c]{@{}l@{}}Oracle\end{tabular} & \begin{tabular}[c]{@{}l@{}}Beam\end{tabular} & \begin{tabular}[c]{@{}l@{}}Original\end{tabular} & \begin{tabular}[c]{@{}l@{}}FactRanker\end{tabular} & \begin{tabular}[c]{@{}l@{}}w/o RC\end{tabular} \\ \midrule
GOLD & No evidence of acute cardiopulmonary process. & - & - & - & - & - & - & \\ \midrule
\multirow{10}{*}{CANDIDATE} & No evidence of acute cardiopulmonary process. & 100 & \textbf{1} & 2 & 7 & \textbf{1} & 2 & \\ 
& No acute cardiopulmonary process.  & 100 & 2 & 4 & 4 & 2 & 4  \\ 
 & \begin{tabular}[c]{@{}l@{}}No evidence of acute intrathoracic process.\end{tabular} & 66.67  & 3 & 3 &  2 & 3 & 3 \\
 & No acute intrathoracic process. & 66.67 & 4& 5& 8 & 4 & 5\\
  & No evidence of acute cardiopulmonary abnormality. & 66.67 & 5 & 7  & 3 & 5 & 7\\ 
  & No evidence of acute process.  & 40.0 & 6 & 9 & 9 & 6 & 9   \\
 & Unremarkable chest radiographic examination. & 0.0 & 7 & \textbf{1}  & 5 & 7 & \textbf{1}  \\ 
 & Normal chest radiographic examination. & 0.0  & 8 & 6 & 10 & 8 & 6   \\ 
 & Unremarkable chest x-ray. & 0.0 & 9 & 8 & 6 & 9 & 8 \\
 & Unremarkable chest examination. & 0.0 & 10  & 10 & \textbf{1} & 10 & 10 \\ 
  \bottomrule
\end{tabular}
\caption{The ranking example of all decoding methods.}
\label{tab:ranking-example}
\end{table*}
\begin{table}[htb]
\begin{tabular}{@{}l|l|l@{}}
\toprule
\begin{tabular}[c]{@{}l@{}}DECODING\\ METHODS\end{tabular} & RadMRR & RadGraph \\ \midrule
Beam Search                                                & 51.8   & 44.3     \\ \midrule
Oracle                                                     & 99.9   & 60.1     \\ \midrule
OriginalReranker                                           & 44.0   & 38.2     \\
FactReranker                                               & 72.6   & 53.2         \\
FactReranker(w/o RC)                                        & 50.4   & 45.0     \\
\bottomrule
\end{tabular}
\caption{The ranking results of all decoding methods. We report the RadMRR and RadGraph scores.}
\label{tab:RadMRR}
\end{table}
On the MIMIC-CXR dataset, our method shows a significant improvement compared with other methods in the factual consistency metric RadGraph score and $F_1$CheXbert.
This demonstrates the superiority of our proposed method in enhancing factual consistency for reranking and selecting the optimal target summary according to the estimated target RadGraph score rather than the log probability.
FactReranker also yields the highest ROUGE-1, ROUGE-2, and ROUGE-L scores among all methods, indicating that the improvement of the factual consistency would also benefit the fluency of the generated summary for generating accurate key entity words and semantic relations.
Specifically, when compared with the SOTA factual-guided summarizer RadGraph-RL,
our method yields a higher improvement than RadGraph-RL over non-factual-guided methods.
It proves the essential for explicitly capturing medical facts in the target summary to select the optimal summary, rather than directly optimizing the factuality score with RL and but select the summary based on the log probability.
Our method outperforms the BART backbone model which selects the top-1 candidate with the beam search decoding by $8\%$ in RadGraph score, which further proves the effectiveness of the factual-guided reranker.
Both our method and RadGraph-RL outperform non-factual-guided methods such as GraphSum and WGSUM on RadGraph score, although they have shown more powerful ability than traditional methods like PGN and TRANSEXT on Rouge score.

As for the OpenI dataset, our method also presents a competitive performance in ROUGE-1, ROUGE-2, and ROUGE-L metrics compared with existing methods.
It has been reported that existing factual consistency metrics such as the RadGraph score and $F_1$CheXbert are ineffective for OpenI since the extraction method is trained on MIMIC-CXR.
Therefore all previous methods and our method don't report the factual consistency results on this dataset.
Compared with the backbone method BART, our method shows a slight improvement in ROUGE scores, which indicates again that the RadGraph score is not effective in this dataset since the extractor would fail to extract correct triplets.
We believe more factual evaluation metrics should be developed in this field, that can be used in various datasets including the OpenI dataset.
\begin{table*}[htb]
\begin{tabular}{@{}l@{}}
\toprule
\begin{tabular}[c]{@{}l@{}}\textbf{Original Text}: The \textcolor{red}{(lungs, ANAT-DP, NA)} are \textcolor{red}{(well, OBS-DP, REL)} \textcolor{red}{(expanded, OBS-DP, REL)} and \textcolor{red}{(clear, OBS-DP, REL)}.\\\textcolor{red}{(Cardiomediastinal, ANAT-DP, NA)} and \textcolor{red}{(hilar, ANAT-DP, NA)} \textcolor{red}{(contours, ANAT-DP, REL)} are \textcolor{red}{(unremarkable, OBS-DP, REL)}. \\There is no \textcolor{red}{(pleural, ANAT-DP, NA)} \textcolor{red}{(effusion, OBS-DA, REL)} or \textcolor{red}{(pneumothorax, OBS-DA, NA)}. \textcolor{red}{(Sternotomy, OBS-DP, REL)} \\ \textcolor{red}{(wires, OBS-DP, NA)} are again noted, with \textcolor{red}{(facture, OBS-DP, REL)} of the \textcolor{red}{(two, OBS-DP, REL)} \textcolor{red}{(upper, OBS-DP, REL)} \\\textcolor{red}{(wires, OBS-DP, NA)} \textcolor{red}{(unchanged, OBS-DP, REL)} from prior exam.\end{tabular}\\
\textbf{Gold Summary}: No evidence of \textcolor{red}{(acute, OBS-DA, REL)} \textcolor{red}{(cardiopulmonary, ANAT-DP, NA)} \textcolor{red}{(process, OBS-DA, REL)}.\\ 
\textbf{FactReranker}: <s>acute [OBS-DA] [REL] [ENT] cardiopulmonary [ANAT-DP] [NA] [ENT] process [OBS-DA] [REL]</s>\\
\textbf{FactReranker(w/o RC)}: <s>[ [ [REL]</s>\\\midrule
\begin{tabular}[c]{@{}l@{}}\textbf{Original Text}: A \textcolor{red}{(hazy, OBS-DP, REL)} \textcolor{red}{(opacity, OBS-DP, REL)} is present in the \textcolor{red}{(right, ANAT-DP, REL)} \textcolor{red}{(lung, ANAT-DP, NA)} \\
which may represent \textcolor{red}{(aspiration, OBS-U, NA)} \textcolor{red}{(pleural, ANAT-DP, NA)} \textcolor{red}{(effusion, OBS-U, REL)} or \textcolor{red}{(hemorrhage, OBS-U, NA)}. \\
\textcolor{red}{(Retrocardiac, ANAT-DP, NA)} \textcolor{red}{(opacity, OBS-DP, REL)} at the \textcolor{red}{(left, ANAT-DP, REL)} \textcolor{red}{(base, ANAT-DP, NA)} is \textcolor{red}{(unchanged, OBS-DP, REL)}. \\
\textcolor{red}{(Moderate, OBS-DP, REL)} \textcolor{red}{(cardiomegaly, OBS-DP, NA)} is \textcolor{red}{(stable, OBS-DP, REL)}. \textcolor{red}{(Slight, OBS-DP, REL)} \textcolor{red}{(prominence, OBS-DP, REL)} \\
of the \textcolor{red}{(pulmonary, ANAT-DP, REL)} \textcolor{red}{(vasculature, ANAT-DP, NA)} with \textcolor{red}{(cephalization, OBS-DP, NA)} and \textcolor{red}{(enlarged, OBS-DP, REL)} \\
\textcolor{red}{(pulmonary, ANAT-DP, REL)} \textcolor{red}{(arteries, ANAT-DP, NA)} are consistent with \textcolor{red}{(mild, OBS-DP, REL)} \textcolor{red}{(pulmonary, ANAT-DP, NA)} \\
\textcolor{red}{(edema, OBS-DP, REL)}. \textcolor{red}{(Tracheostomy, OBS-DP, REL)} \textcolor{red}{(tube, OBS-DP, NA)} is \textcolor{red}{(in place, OBS-DP, REL)}. \\
There are no \textcolor{red}{(displaced, OBS-DA, REL)} \textcolor{red}{(rib, ANAT-DP, NA)} \textcolor{red}{(fractures, OBS-DA, REL)}.\end{tabular}\\
\textbf{Gold Summary}:1. \textcolor{red}{(Hazy, OBS-DP, REL)} \textcolor{red}{(opacity, OBS-DP, REL)} in the \textcolor{red}{(right, ANAT-DP, REL)} \textcolor{red}{(lung, ANAT-DP, NA)} which may \\
represent \textcolor{red}{(aspiration, OBS-U, NA)} versus \textcolor{red}{(pleural, ANAT-DP, NA)} \textcolor{red}{(effusion, OBS-U, REL)} or \textcolor{red}{(hemorrhage, OBS-U, NA)} \\
2. \textcolor{red}{(Mild, OBS-DP, REL)} \textcolor{red}{(pulmonary, ANAT-DP, NA)} \textcolor{red}{(edema, OBS-DP, REL)} \\
3. No \textcolor{red}{(displaced, OBS-DA, REL)} \textcolor{red}{(rib, ANAT-DP, NA)} \textcolor{red}{(fractures, OBS-DA, REL)}\\
\textbf{FactReranker}: <s>Hazy [OBS-DP] [REL] [ENT] opacity [OBS-DP] [REL] [ENT] right [ANAT-DP] [REL] [ENT] \\
lung [ANAT-DP] [NA] [ENT] aspiration [OBS-U] [NA] [ENT] pleural [ANAT-DP] [NA] [ENT] effusion [OBS-U] [REL] [ENT] \\
hemorrhage [OBS-U] [NA] [ENT] Mild [OBS-DP] [REL] [ENT] pulmonary [ANAT-DP] [NA] [ENT] edema [OBS-DP] [REL]</s>\\
\textbf{FactReranker(w/o RC)}: <s></s>\\
\bottomrule
\end{tabular}
\caption{The generated triplet sequences from FactRanker and the ablation FactRanker(w/o RC), along with source documents and gold summaries annotated with RadGraph triplets using red text color.}
\label{tab:triplet-example}
\end{table*}
\begin{table}[htb]
\begin{tabular}{@{}l|clll@{}}
\toprule
MODEL         & \multicolumn{4}{c}{MIMIC-CXR}                                   \\ \midrule
METRIC       & R-1                  & R-2   & R-3   & BertScore \\ \midrule
FactRanker          &\textbf{81.72}&\textbf{74.82}&\textbf{80.52}&\textbf{94.87} \\ 
FactRanker(w/o RC)         &0.10&0.10&0.10&81.48 \\ 
\bottomrule
\end{tabular}
\caption{The ROUGE score and BertScore of FactRanker and FactRanker(w/o RC) on MIMIC-CXR}
\label{tab:triplet-performance}
\end{table}
\subsection{Reranking Performance}
In this section, we further investigate \textbf{RQ1} with a detailed question: 
Does our second-stage reranker possess the ability to rank candidates according to their factual consistency?
To answer it, we conduct experiments on three different ranking methods:
1) \textbf{Beam Search}: it generates 10 candidate sequences and selects the optimal one with the highest log-likelihood;
2) \textbf{Oracle}: it selects the optimal candidate according to its factual consistency with the gold summary;
3) \textbf{OriginalRadRanker}: it first calculates the RadGraph scores of all candidates with the original texts and chooses the one with the highest RadGraph score.
4) \textbf{Ours}:  Similar to OriginalRadRanker, our method calculates the RadGraph scores of all candidates with the predicted factual knowledge graph of the gold summary and chooses the one with the highest RadGraph score.
We compare the results of these methods in Table~\ref{tab:RadMRR}, which utilizes our designed ranking metric RadMRR to measure the ranking ability.
As aforementioned in Section~\ref{sec:radmrr}, the higher RadMRR, the higher position of the optimal candidate in the ranked list of all candidates.
We also present the RadGraph score of the selected summary from different ranking methods.
The results clearly prove the ability of our method to rerank candidates according to their factual relevance to the target summary.
In comparison with Beam search, which is generally adopted by previous methods, our method yields a 21\% higher RadMRR and results in a significantly better RadGraph score, which is competitive with the results of Oracle.
This is also proved when Oracle has a 20\% improvement over Beam Search, which is almost two times larger than the improvement in the previously proposed research.
Our method also outperforms OrigianlRadRanker that utilizes the source document with no regard for the semantic gap in the medical facts between the source document and the target summary.
In fact, OriginalRadRanker has the worst ranking performance among all ranking methods, which indicates a significant difference between the medical facts of source texts and target summaries.
In Table~\ref{tab:ranking-example}, we additionally provide an example of a list of candidates and their corresponding positions generated by each ranking method.
Compared with existing ranking methods, our method has the same ranking result with that of the Oracle. It can generally assign high-factual consistent candidates to higher positions and otherwise lower positions, leading to a more robust and faithful summarization process.
\subsection{Ablation Study}
To answer \textbf{RQ2} and attribute the ranking ability of our reranker,
we further perform experiments on our method along with \textbf{w/o RC} where we directly adopt the first-stage summarizer to generate the RadGraph triplets of the gold summary without further fine-tuning.
First, as shown in Table~\ref{tab:RadMRR}, our method of fine-tuning the RadGraph generation presents better ranking performance and higher factual correctness, which indicates the importance of continuous training of the existing summarizer for RadGraph generation.
Compared with our method, \textbf{w/o RC} presents a 22\% decrease in RadMRR and 8\% in RadGraph score, which is almost the same performance as the beam search method.
This proves that the improvement of our method over beam search mostly attributes to the RadGraph generation which explicitly models the medical facts in the target summary, rather than the first-stage summarizer as the backbone model.
As shown in Table~\ref{tab:RadMRR} and Table~\ref{tab:ranking-example}, with wrong medical facts in the gold summary, \textbf{w/o RC} presents the comparable ranking performance with the first-stage summarizer based on the beam search.
\subsection{Case Study}
In this section, we further investigate the performance of our triplet generation model on generating the medical facts of the gold summary to answer \textbf{RQ3}.
In Table~\ref{tab:triplet-example}, we present the generated sequences of our method and the ablation, along with the ground truth RadGraph of the original text and its gold summary.
The first example in the first row clearly shows that the RadGraph generation model in our reranker can generate the correct RadGraph triplets of the gold summary, that covers the groud truth medical facts in the gold summary.
Even for the difficult sample as shown in the second row, the fine-tuned RadGraph generation method in our reranker can still predict the RadGraph triplets, that has high overlap with the ground truth RadGraph triplets of the gold summary.
In contrast, directly applying summarization models without fine-tuning would result in sequences with broken triplets.
Consequently, the model would select the first candidate as the beam search since all candidates have the same score due to an empty generated triplet set.
Moreover, we adopt several metrics to evaluate the generated sequences for triplet generation, including ROUGE and BertScore.
As shown in Table~\ref{tab:triplet-performance}, FactRanker with fine-tuning on the RadGraph constrcution yields a relatively high ROUGE score and BertScore, indicating the ability of our generation model to generate faithful RadGraph triplets of the target summary.

\section{Conclusion}
In this paper, we present the first second-state approach FactReranker for faithful radiology report summarization that prioritizes factual consistency of the generated summary. We propose the fact-guided reranker by extracting medical facts based on the RadGrap and incorporating them into the reranking process. We decompose the fact-guided reranker into the RadGraph generation and factual scorer, which allows the reranker to precisely estimate the RadGraph scores of candidate summaries and choose the optimal summary according to the predicted RadGraph of the gold summary. Our experimental results show that FactReranker outperforms existing state-of-the-art radiology report summarization methods and provides a promising solution to the challenge of ensuring factual accuracy in this task. 
For future work, we will explore dealing with limitations in our method: 1) In the first stage, we only use the BART backbone model with the beam search decoding method for generating candidates. We will further investigate and analyze the influence of different backbone models and decoding strategies on the performance
2) In the second stage, we only adopt the RadGraph triplets from both the source document and target summary. We will further investigate using the natural language texts to further improve the generation performance base don the prompt learning.
Moreover, we will further explore developing a reranker that can learn to select the optimal candidate based on multiple metrics.

\section*{Acknowledgment}
This work was supported by the National Library of Medicine grant 4R00LM013001, NSF CAREER Award No. 2145640, and Amazon Research Award. We would like to thank Jinpeng Hu for providing the processed dataset and useful feedback for the paper. We also would like to thank Jimin Huang for his useful suggestions on improving the paper.
\bibliographystyle{ACM-Reference-Format}
\bibliography{custom}

\appendix

\end{document}